\pdfoutput=1

\documentclass[11pt]{article}

\usepackage{acl}

\usepackage{times}
\usepackage{latexsym}

\usepackage{booktabs}
\usepackage{graphicx}
\usepackage{subfiles}
\usepackage{cmap}
\usepackage{tabularx} 
\usepackage{comment} 

\usepackage[T1]{fontenc}

\usepackage[utf8]{inputenc}

\usepackage{microtype}

\title{5q032e@SMM4H'22: Transformer-based classification of premise in tweets related to COVID-19}

\author{Vadim Porvatov \\
  Sberbank / Moscow 117997, Russia \\
  \texttt{eighonet@gmail.com} \\\And
  Natalia Semenova \\
  AIRI / Moscow 105064, Russia \\
  Sberbank / Moscow 117997, Russia \\
  \texttt{semenova.bnl@gmail.com} \\}

\begin{document}
\maketitle
\begin{abstract}
Automation of social network data assessment is one of the classic challenges of natural language processing. During the COVID-19 pandemic, mining people's stances from public messages have become crucial regarding understanding attitudes towards health orders. In this paper, the authors propose the predictive model based on transformer architecture to classify the presence of premise in Twitter texts. This work is completed as part of the Social Media
Mining for Health (SMM4H) Workshop 2022. We explored modern transformer-based classifiers in order to construct the pipeline efficiently capturing tweets semantics. Our experiments on a Twitter dataset showed that RoBERTa-large is superior to the other transformer models in the case of the premise prediction task. The model achieved competitive performance with respect to ROC AUC value 0.807, and 0.7648 for the F1 score.
\end{abstract}

\section{Introduction}

Modern natural language processing methods emerged as tools of outstanding performance in many classic machine learning tasks. Along with their other achievements, the introduction of transformer architecture~\citep{vaswani2017attention} allowed to develop of domain-specific models in the interlingual transfer of social media texts~\citep{miftahutdinov2020biomedical}, identification of drug similarity~\citep{tutubalina2017using}, and detection of adverse drug effects~\citep{sakhovskiy2021kfu}.

One of the prospective domains of language model development is an automatic extraction and further assessment of insights gained from Twitter texts~\citep{pamungkas2019stance}. In addition, stance and premise classification tasks are frequently interpreted as critical challenges in social media text analysis~\cite{bar-haim-etal-2017-stance,premise_article}.

\textbf{Contribution}. In this paper, we extensively evaluate premise classification approaches and partially explore dependencies between configurations of the considered models and their performance. 

\section{Data}

\begin{figure}[!t]
\includegraphics[width=219pt]{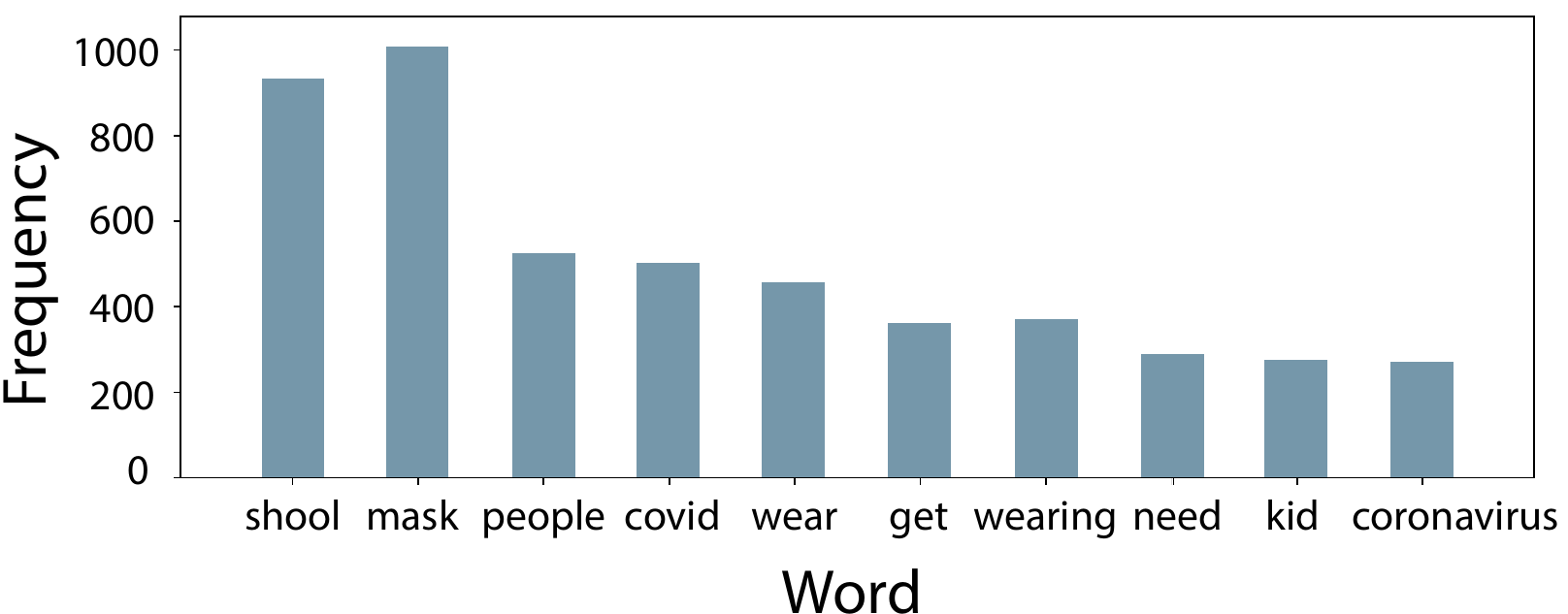}
\caption{Top-10 frequent words in the dataset.} \label{projection}
\label{data:freqs}
\end{figure}

Available labeled data~\citep{davydova2022dataset} includes 4155
tweets divided into train and test samples in a ratio of 17:3. Regarding the premise classification, the dataset contains a subset of 2445 tweets with a positive label and 1710 tweets with a negative label. As an additional metadata, there are 1402 tweets tagged as \textit{stay at home orders}, 1526 related to the \textit{face masks}, and 1227 tweets marked as opinions about \textit{school closures}. Established text data could be assessed from the perspective of trending word frequencies, Figure~\ref{data:freqs}.

\begin{table*}[t]
\small
\centering
\begin{tabularx}{410pt}{X|X|XXX|XXX}
\toprule
\toprule
\multicolumn{2}{X|}{ Split} & \multicolumn{3}{X|}{Train} & \multicolumn{3}{X}{Test} \\
\midrule
\multicolumn{2}{c|}{} & Accuracy & F1-score & ROC AUC & Accuracy & F1-score & ROC AUC  \\
\midrule
\multicolumn{2}{l|}{Random} &0.4986 & 0.5592 & 0.5014 & 0.4959  & 0.4302 & 0.5016 \\
\midrule
\multicolumn{2}{l|}{BERT (base, uncased)} & 0.9446 & 0.9268 & 0.9392 & 0.7947  & 0.7185 & 0.7793 \\
\multicolumn{2}{l|}{BERT (base, uncased, ml)} & 0.913 & 0.889 & 0.9005 & 0.7813  & 0.7385 & 0.7742 \\
\midrule
\multicolumn{2}{l|}{AlBERTv2 (base)} & 0.9781 & 0.9708 & 0.9758 & 0.7746 & 0.6853 & 0.7581 \\
\multicolumn{2}{l|}{AlBERTv2 (xlarge)} & 0.9215 & 0.8992 & 0.9126 & 0.7496 & 0.6681 & 0.7314 \\
\midrule
\multicolumn{2}{l|}{DeBERTa (base)} & 0.9194 & 0.8995 & 0.9095 & 0.813 & 0.7607 & 0.799 \\
\midrule
\multicolumn{2}{l|}{Longformer (large)} & 0.9758 & 0.9681 & 0.9721 & 0.8097 & 0.7522 & 0.7956 \\
\midrule
\multicolumn{2}{l|}{RemBERT} & \textbf{0.9885} &\textbf{ 0.9846} & \textbf{0.9871 } & 0.7997 & 0.7309 & 0.7842 \\
\midrule
\multicolumn{2}{l|}{RoBERTa (base)} & 0.9213 & 0.9024 & 0.9118 & 0.7997 & 0.7521 & 0.7882 \\
\multicolumn{2}{l|}{RoBERTa (large)} & 0.9837 & 0.9761 & 0.9795 & \textbf{0.8214} & \textbf{0.7648} & \textbf{0.807}\\

\bottomrule
\bottomrule
\end{tabularx}
\caption{Evaluation of methods for a premise classification task.}
\label{results_table_1}
\end{table*}

\section{Method}

In order to reach the best score, we performed analysis of the state-of-the-art architectures for text classification and selected the following models: BERT~\citep{bert}, RoBERTa~\citep{roberta}, AlBERT~\citep{albert}, DeBERTa~\citep{deberta}, RemBERT~\citep{rembert}, and Longformer~\citep{Beltagy2020Longformer}. Generally, these models encode each tweet to the fixed-size vector and further apply the classifier layers in an end-to-end manner. The output activation function (softmax) converts the hidden representation of the text to the desired class probabilities which are further used during the computation of binary cross entropy loss function:
\begin{equation}
-\frac{1}{n} \sum_{i=1}^ny_i \cdot \log \hat{y}_i+\left(1-y_i\right) \cdot \log \left(1-\hat{y}_i\right),
\end{equation}
where $n$ is the number of samples, $y_i$ is the true label of a tweet, and $\hat{y}_i$ is the predicted one.

Relatively small train data encouraged us to use pre-trained variations of each considered model.  As long as tweets comprise of divergent content, it is required to inspect the dataset at the preprocessing stage carefully. To move further with model training, we need to handle the presence of nametags, hashtags, emoticons, and other additional symbols.
We perform processing procedures with the help of \textit{tweet-preprocessor}\footnote{https://pypi.org/project/tweet-preprocessor/} package. First, we remove abundant text pieces (e. g., user mentions) as well as replace web pages links and hashtags with placeholders. Before tokenization, we convert cased words to uncased ones.

\section{Experiments}

\begin{figure}[!t]
\includegraphics[width=219pt]{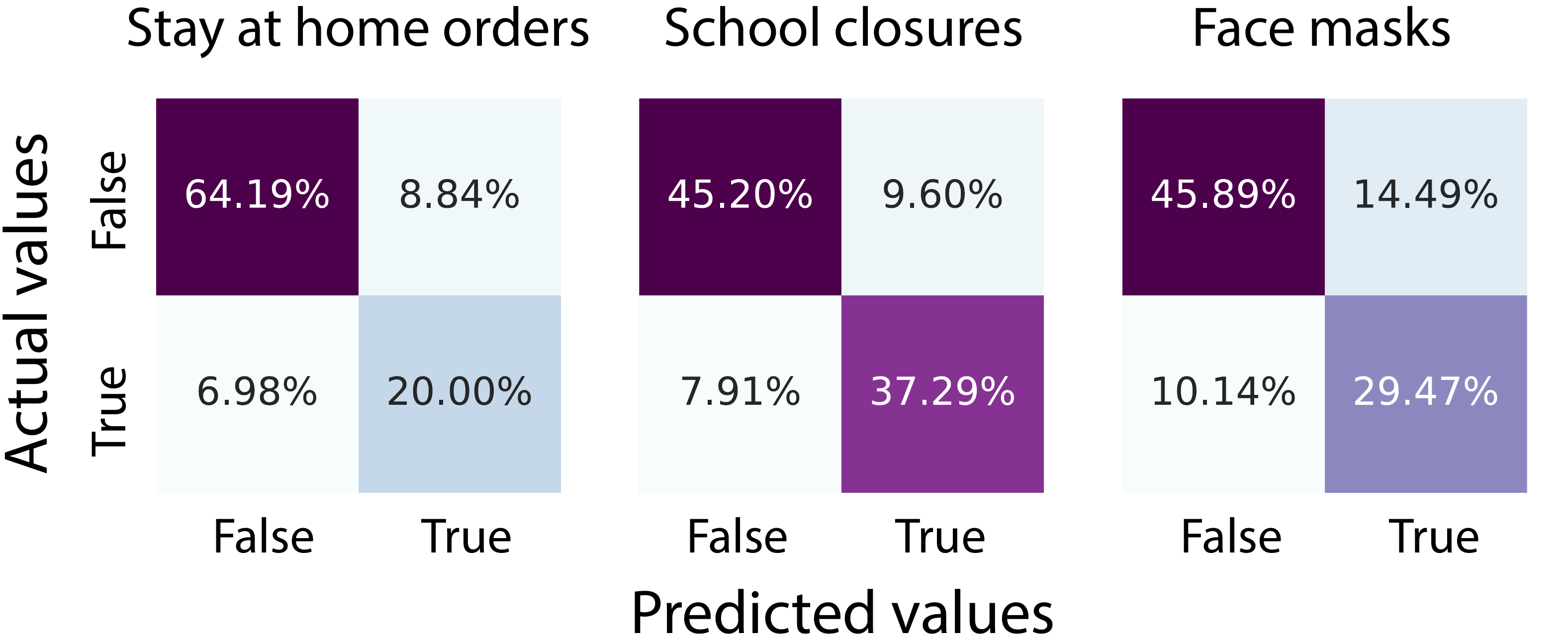}
\caption{Confusion matrices of RoBERTa-large regarding the different tweets categories of test data.} 
\label{results:cms}
\end{figure}


We measure the performance of the models on the main task via three commonly used metrics for the binary classification: ROC AUC, Accuracy, and F1-score.

As the optimizer for the selected models, we used AdamW~\citep{Loshchilov2019DecoupledWD}. The models were trained along the 20 epochs with learning rates varying from 0.001 to 0.00001 and batch sizes from the range [4, 8, 16, 32, 48]. Experiments were done with 3 Tesla V100 GPUs and 512 Gb of RAM. The training time of the models lies in the interval from 3.5 hours up to 5 hours.

For the premise classification, the best performance was achieved by the large RoBERTa model, Table~\ref{results_table_1}. Obtained metrics are tangibly dissimilar from the other architectures despite RemBERT which suffers from overfitting and thus converges to the better values in the training sample. Detailed analysis of RoBERTa-large performance on different tweets categories is given in Figure~\ref{results:cms}.

To ensure the statistical significance of applied preprocessing procedures, we leverage the Mann–Whitney U test regarding the null hypothesis that the F1 scores obtained on the initial and preprocessed tweets belong to the same distribution. We rejected the null hypothesis with a p-value $\leq$ 0.05.

\section{Conclusion}

In this work, we have explored the application of different transformer models to the task of premise classification. We extensively evaluated BERT variants and obtained the best architecture during the computational experiments. In future work, we intend to focus on the ensembling methods applied to the presented task.

\bibliography{acl2020}

\begin{thebibliography}{15}
\expandafter\ifx\csname natexlab\endcsname\relax\def\natexlab#1{#1}\fi

\bibitem[{Bar-Haim et~al.(2017)Bar-Haim, Bhattacharya, Dinuzzo, Saha, and
  Slonim}]{bar-haim-etal-2017-stance}
Roy Bar-Haim, Indrajit Bhattacharya, Francesco Dinuzzo, Amrita Saha, and Noam
  Slonim. 2017.
\newblock \href {https://aclanthology.org/E17-1024} {Stance classification of
  context-dependent claims}.
\newblock In \emph{Proceedings of the 15th Conference of the {E}uropean Chapter
  of the Association for Computational Linguistics: Volume 1, Long Papers},
  pages 251--261, Valencia, Spain. Association for Computational Linguistics.

\bibitem[{Beltagy et~al.(2020)Beltagy, Peters, and
  Cohan}]{Beltagy2020Longformer}
Iz~Beltagy, Matthew~E. Peters, and Arman Cohan. 2020.
\newblock Longformer: The long-document transformer.
\newblock \emph{arXiv:2004.05150}.

\bibitem[{Chung et~al.(2020)Chung, Fevry, Tsai, Johnson, and Ruder}]{rembert}
Hyung~Won Chung, Thibault Fevry, Henry Tsai, Melvin Johnson, and Sebastian
  Ruder. 2020.
\newblock Rethinking embedding coupling in pre-trained language models.
\newblock In \emph{International Conference on Learning Representations}.

\bibitem[{Davydova and Tutubalina(2022)}]{davydova2022dataset}
Vera Davydova and Elena Tutubalina. 2022.
\newblock Smm4h 2022 task 2: Dataset for stance and premise detection in tweets
  about health mandates related to covid-19.
\newblock In \emph{Proceedings of the Seventh Social Media Mining for Health
  Applications (SMM4H) Workshop \& Shared Task}, pages~--.

\bibitem[{Go et~al.(2009)Go, Bhayani, and Huang}]{premise_article}
Alec Go, Richa Bhayani, and Lei Huang. 2009.
\newblock Twitter sentiment classification using distant supervision.
\newblock \emph{Processing}, 150.

\bibitem[{He et~al.(2020)He, Liu, Gao, and Chen}]{deberta}
Pengcheng He, Xiaodong Liu, Jianfeng Gao, and Weizhu Chen. 2020.
\newblock Deberta: Decoding-enhanced bert with disentangled attention.
\newblock In \emph{International Conference on Learning Representations}.

\bibitem[{Kenton and Toutanova(2019)}]{bert}
Jacob Devlin Ming-Wei~Chang Kenton and Lee~Kristina Toutanova. 2019.
\newblock Bert: Pre-training of deep bidirectional transformers for language
  understanding.
\newblock In \emph{Proceedings of NAACL-HLT}, pages 4171--4186.

\bibitem[{Lan et~al.(2019)Lan, Chen, Goodman, Gimpel, Sharma, and
  Soricut}]{albert}
Zhenzhong Lan, Mingda Chen, Sebastian Goodman, Kevin Gimpel, Piyush Sharma, and
  Radu Soricut. 2019.
\newblock Albert: A lite bert for self-supervised learning of language
  representations.
\newblock In \emph{International Conference on Learning Representations}.

\bibitem[{Liu et~al.(2019)Liu, Ott, Goyal, Du, Joshi, Chen, Levy, Lewis,
  Zettlemoyer, and Stoyanov}]{roberta}
Yinhan Liu, Myle Ott, Naman Goyal, Jingfei Du, Mandar Joshi, Danqi Chen, Omer
  Levy, Mike Lewis, Luke Zettlemoyer, and Veselin Stoyanov. 2019.
\newblock Roberta: A robustly optimized bert pretraining approach.

\bibitem[{Loshchilov and Hutter(2019)}]{Loshchilov2019DecoupledWD}
Ilya Loshchilov and Frank Hutter. 2019.
\newblock Decoupled weight decay regularization.
\newblock In \emph{International Conference on Learning Representations}.

\bibitem[{Miftahutdinov et~al.(2020)Miftahutdinov, Alimova, and
  Tutubalina}]{miftahutdinov2020biomedical}
Zulfat Miftahutdinov, Ilseyar Alimova, and Elena Tutubalina. 2020.
\newblock On biomedical named entity recognition: experiments in interlingual
  transfer for clinical and social media texts.
\newblock In \emph{European Conference on Information Retrieval}, pages
  281--288. Springer.

\bibitem[{Pamungkas et~al.(2019)Pamungkas, Basile, and
  Patti}]{pamungkas2019stance}
EW~Pamungkas, V~Basile, and V~Patti. 2019.
\newblock Stance classification for rumour analysis in twitter: Exploiting
  affective information and conversation structure.
\newblock In \emph{2nd International Workshop on Rumours and Deception in
  Social Media (RDSM 2018)}, volume 2482, pages 1--7. CEUR-WS.

\bibitem[{Sakhovskiy et~al.(2021)Sakhovskiy, Miftahutdinov, and
  Tutubalina}]{sakhovskiy2021kfu}
Andrey Sakhovskiy, Zulfat Miftahutdinov, and Elena Tutubalina. 2021.
\newblock Kfu nlp team at smm4h 2021 tasks: Cross-lingual and cross-modal
  bert-based models for adverse drug effects.
\newblock In \emph{Proceedings of the Sixth Social Media Mining for Health
  (SMM4H) Workshop and Shared Task}, pages 39--43.

\bibitem[{Tutubalina et~al.(2017)Tutubalina, Miftahutdinov, Nugmanov,
  Madzhidov, Nikolenko, Alimova, and Tropsha}]{tutubalina2017using}
EV~Tutubalina, Z~Sh Miftahutdinov, RI~Nugmanov, TI~Madzhidov, SI~Nikolenko,
  IS~Alimova, and AE~Tropsha. 2017.
\newblock Using semantic analysis of texts for the identification of drugs with
  similar therapeutic effects.
\newblock \emph{Russian Chemical Bulletin}, 66(11):2180--2189.

\bibitem[{Vaswani et~al.(2017)Vaswani, Shazeer, Parmar, Uszkoreit, Jones,
  Gomez, Kaiser, and Polosukhin}]{vaswani2017attention}
Ashish Vaswani, Noam Shazeer, Niki Parmar, Jakob Uszkoreit, Llion Jones,
  Aidan~N Gomez, {\L}ukasz Kaiser, and Illia Polosukhin. 2017.
\newblock Attention is all you need.
\newblock \emph{Advances in neural information processing systems}, 30.

\end{thebibliography}
\bibliographystyle{acl_natbib}

\end{document}